\documentclass[lettersize,journal]{IEEEtran}
\usepackage{amsmath,amssymb,amsthm}
\usepackage{graphicx} 
\usepackage{epsfig}
\usepackage{amsfonts}
\usepackage{amstext}
\usepackage{algorithm}
\usepackage[noend]{algpseudocode}
\usepackage[caption=false,font=normalsize,labelfont=sf,textfont=sf]{subfig}
\usepackage{bm}

\newcommand{\done}{\hspace*{\fill} $\Box$}
\newcommand{\RE}{I\!\!R}

\newtheorem{theorem}{Theorem}
\newtheorem{remark}{Remark}
\newtheorem{corollary}{Corollary}

\newtheorem{lemma}{Lemma}
\newtheorem{example}{Example}

\onecolumn

\begin{document}

\title{\LARGE \bf Analysis and Design of Quadratic Neural Networks for Regression, Classification, and Lyapunov Control of Dynamical Systems}

\author{Luis Rodrigues$^1$ and Sidney Givigi$^2$\\
$^1$Department of Electrical and Computer Engineering\\
 Concordia University, 1515 St. Catherine Street, Montr\'eal, QC H3G 2W1, Canada\\
$^2$ School of Computing\\
Queens University, 557 Goodwin Hall, Kingston, Ontario, K7L2N8, Canada\\}

\markboth{Preprint Version, May 2022}%
{Shell \MakeLowercase{\textit{et al.}}: A Sample Article Using IEEEtran.cls for IEEE Journals}


\maketitle

\begin{abstract}
This paper addresses the analysis and design of quadratic neural networks, which have been recently introduced in the literature, and their applications to regression, classification, system identification and control of dynamical systems.
These networks offer several advantages, the most important of which are the fact that the architecture is a by-product of the design and is not determined a-priori, their training can be done by solving a convex optimization problem so that the global optimum of the weights is achieved, and the input-output mapping can be expressed analytically by a quadratic form.
It also appears from several examples that these networks work extremely well using only a small fraction of the training data.
The results in the paper cast regression, classification, system identification, stability and control design as convex optimization problems, which can be solved efficiently with polynomial-time algorithms to a global optimum. Several examples will show the effectiveness of quadratic neural networks in applications.
\end{abstract}


\section{Introduction}
The first mathematical model of an artificial neuron was proposed in 1943 by McCulloch and Pitts \cite{McCullochPitts1943}.
Since then, artificial neural networks have been an active area of research with applications such as image recognition, natural language processing, and signal processing to name a few.
However, their widespread use did not yet reach the same proportion in safety-critical applications of control systems, such as autonomous vehicles, because of the lack of theoretical guarantees on safety, stability, and performance, which are of prime concern for such applications.
In particular, formal results on Lyapunov stability of a system in feedback with a neural network controller are very scarce and only started appearing very recently in the literature \cite{Pavonetal2021}.
Another current issue with most artificial neural networks is that before the training of the network can be performed one must decide on the network architecture.
This is most often a difficult trial-and-error task that is heavily dependent on the application, although upper bounds on the number of neurons in feedforward networks needed to learn a given number of data points were determined in reference \cite{Huang1998}.
Architectures in which the activation functions are different in different layers have been proposed for efficient approximation of high dimensional functions in \cite{Cheridito2021}.
The architecture redundancy for pixel level estimation was recently addressed in reference \cite{Luo2021}.
Once an architecture has been chosen, another important issue is that the training of the neural network weights does not usually guarantee that the global optimum value is achieved.
An additional difficulty in the analysis of neural networks is the fact that the mapping between the input and output cannot be written as a concise analytical expression for a general input.
This makes it difficult to estimate the robustness of the output of the neural network for small changes in the input, which is typically done by computing a Lipschitz continuity constant \cite{Fazlyab2019,Allgoweretall2022}.
Moreover, the best current results of training neural networks using deep learning \cite{Bengioetal2016} typically require an extremely large amount of data, so-called {\em big-data}.

The training of quadratic neural networks (QNN) was recently shown in reference \cite{BartanPilanci2021} to be a convex optimization program with guarantees of achieving the global optimum value of the weights.
Additionally, quadratic neural networks also offer advantages relative to many other  issues mentioned in the previous paragraph. In particular, the input and output are related by a quadratic form and the architecture of the network is a by-product of the training itself.
This comes at the expense of only having one hidden layer, although extensions to more hidden layers have been proposed in reference \cite{BartanPilanci2021}.

The main focus of this paper is to perform a theoretical analysis and to provide several simulation results for quadratic neural networks applied to regression, classification, system identification, and control problems.
The paper is organized as follows. Section \ref{quadraticnetworks} will review quadratic neural networks and provide three results on their relationship with quadratic forms. Then regression, classification, and system identification will be addressed in section \ref{regressionclassification}. This is followed by control of dynamical systems in section \ref{controlsystems}.
After presenting several examples in section \ref{examples} the paper ends with a final discussion and the conclusions.

\section{Quadratic Neural Networks}\label{quadraticnetworks}
\begin{figure}[t] 
\centerline{ \resizebox{95mm}{!}{\includegraphics{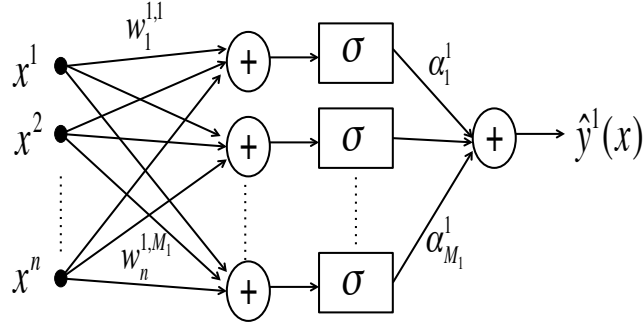}}}
\caption{Single output quadratic neural network}
\label{neuralnetwork}
\end{figure}

The structure of a quadratic feedforward neural network with one output as proposed in \cite{BartanPilanci2021} is shown in figure \ref{neuralnetwork}.
The architecture of the single output neural network can be repeated for the case of $p$ outputs.
The network proposed in \cite{BartanPilanci2021} is constrained to have a single hidden layer with $M=\sum_{k=1}^pM_k$ neurons.
Each output is connected to $M_k$ neurons in the hidden layer.
The value of each network output is obtained by a linear combination of the outputs of the hidden layer neurons. 
The mathematical expression of each output $k=1,\ldots,p,$ is
\begin{equation}\label{networkoutput}
\hat y^k(x) = \hat f^k(x) = \sum_{j=1}^{M_k}\sigma\left(x^Tw^{k,j}\right)\alpha_j^k
\end{equation}
where the activation function is quadratic and is written as
\begin{equation}\label{activationfunction}
\sigma(z) = az^2 + bz + c
\end{equation}
where $a\neq 0, b, c,$ are pre-defined constants that parameterize the quadratic activation function.
The weights $w$ connect the input $x\in\RE^n$ to each neuron in the hidden layer.
The notation of equation (\ref{networkoutput}) where the weights leaving the input are denoted by $w$ and the weights leaving the hidden layer are denoted by $\alpha$ will be used throughout the paper. For the weights we will use superscript indices to indicate the target  and subscript indices to indicate the source of each connection.
For example, to denote the weight connecting the first input component to the second hidden neuron of the third output we use $w_1^{3,2}$.
The desired (label) outputs will be denoted by $y$ and the actual outputs of the network will be denoted by $\hat y$.
The norm-$1$ and norm-$2$ of a vector  are $\|v\|_1=\sum_i|v_i|$ and $\|v\|_2^2=\sum_iv_i^2$.

One of the advantages of quadratic neural networks is that the training of the weights can be done by solving a convex optimization problem, which guarantees convergence to the global optimum.
Following \cite{BartanPilanci2021} it will be assumed that the weights $w^{k,j}$ are normalized to have unit norm.
Additionally, a regularization term on the $1$-norm of $\alpha_j$ will be included in the objective function for the network training.
Using a convex loss function $l(\cdot)$, the original (primal) non-convex training problem for a quadratic network where all hidden neurons are connected to all outputs (using  all weights $w^j$ instead of $w^{k,j}$ for each output $\hat y^k$) is defined as \cite{BartanPilanci2021}
\begin{equation}\label{nonconvextraining}
\begin{array}{l}
\min\limits_{w^{j},\alpha_j}  l(\hat y-y) + \beta\sum_{i=1}^M\|\alpha_i\|_1 \\
\mbox {s.t.}~\hat y^k = \sum_{j=1}^{M}\sigma\left(x^Tw^{j}\right)\alpha_j^k,\\
~~~~\|w^{j}\|_2=1,~k=1,\ldots,p,~j=1,\ldots,M,\\
\end{array}
\end{equation}
for fixed $a\neq 0, b, c,$ and a fixed regularization coefficient $\beta\ge 0$.
Following reference \cite{BartanPilanci2021}, unless otherwise stated the parameters $a,b,c,$ will be selected to approximate in the least squares sense the function $ReLU(z)=\max(0,z)$ in the interval $[-5,5]$ and are equal to $a=0.0937, b=0.5, c=0.4688$.
The following result from reference \cite{BartanPilanci2021} recasts the training as an equivalent convex optimization problem.
\vspace{10pt}
\begin{lemma}\cite{BartanPilanci2021}
Given fixed $a\neq 0, b, c,$ and a fixed regularization coefficient $\beta\ge 0$, the solution of the convex problem that is dual to (\ref{nonconvextraining}) and is formulated as
\begin{equation}\label{convextraining}
\begin{array}{l}
\min\limits_{}  l(\hat y-y) + \beta\sum_{k=1}^p\left(Z^{k,4}_+ + Z^{k,4}_-\right) \\
\mbox {s.t.}~\\
\hat y_i^k = \bar x_i^T\left[
\begin{array}{cc}
a\left(Z^{k,1}_+-Z^{k,1}_-\right) & \frac{b}{2}\left(Z^{k,2}_+-Z^{k,2}_-\right)\\
\frac{b}{2}\left(Z^{k,2}_+-Z^{k,2}_-\right)^T & c{\rm \bf Trace}\left(Z^{k,1}_+-Z^{k,1}_-\right)
\end{array}
\right]
\bar x_i\\
Z^{k,4}_+ = {\rm \bf Trace}\left(Z^{k,1}_+\right), Z^{k,4}_- = {\rm \bf Trace}\left(Z^{k,1}_-\right),\\
Z^k_+ =\left[\begin{array}{cc}
		Z^{k,1}_+ & Z^{k,2}_+\\
		\left(Z^{k,2}_+\right)^T& Z^{k,4}_+
		\end{array}
		\right],
		~Z^k_-=\left[\begin{array}{cc}
		Z^{k,1}_- & Z^{k,2}_-\\
	        \left(Z^{k,2}_-\right)^T & Z^{k,4}_-
		\end{array}
		\right],\\
Z^k_+\ge 0,~Z^k_-\ge 0,~\bar x_i^T=[x_i^T~~1],~k=1,\ldots,p,~i=1,\ldots N, 
\end{array}
\end{equation}
where $l(\cdot)$ is a convex loss function, provides a global optimal solution for the parameters $Z^k_+,Z^k_-\in\RE^{(n+1)\times(n+1)},$ for $k=1,\ldots,p,$ when the number of neurons satisfies $M\ge M_*$ with
\begin{equation}
M_* = \sum_{k=1}^p\left[rank\left(Z^{k*}_+\right) + rank\left(Z^{k*}_-\right)\right],
\end{equation}
where $Z^{k*}_+$ and $Z^{k*}_-$ for $k=1,\ldots,p$, are the solution of the optimization problem (\ref{convextraining}) given $N$ input data vectors $x_i\in\RE^n$ with corresponding labels $y_i\in\RE^{p}$.
Moreover, the optimal value of the solutions of problems (\ref{nonconvextraining}) and (\ref{convextraining}) are the same and therefore the duality gap is zero. \done
\end{lemma}
\vspace{10pt}
A clear advantage of quadratic neural networks is that the optimal network weights and architecture can be extracted from $Z^{k*}_+$ and $Z^{k*}_-$ using algorithm \ref{neuraldecomposition} from reference \cite{BartanPilanci2021} where
\begin{equation}\label{Gmatrix}
G=\left[
\begin{array}{cc}
I_n & 0\\
0 & -1
\end{array}
\right]
\end{equation}
and $I_n$ is the identity matrix of order $n$.

\begin{minipage}[c]{0.20 \textwidth}
\end{minipage}
\hfill
\begin{minipage}[c]{0.5 \textwidth}
\centering
\noindent
\begin{algorithm}[H]
\caption{Neural Decomposition with Tolerance $\bm{tol}$~\cite{BartanPilanci2021}}\label{neuraldecomposition}
\begin{algorithmic}[1]
\Function{NeuralDecomposition~}{$\bm{Z^*},\bm{tol}$}
	\State Compute rank-1 decomposition $\bm{Z^*}=\sum_{j=1}^rp_jp_j^T$

     	using engenvector decomposition keeping all
	
	eigenvectors with eigenvalues  greater than $\bm{tol}$.
	\State Create a list of vectors $\bm{p=\{p_1,\ldots,p_r\}}$ and

	a void output list $\bm{V}$
	\For {$k=1,\ldots,r-1$}

	$~~~p_1=p_k$
	\If {$p_1^TGp_1=0$} $v^*=p_1$
	\Else {~find $j\in\{k+1,\ldots,r\}$: $\left(p_1^TGp_1\right)\left(p_j^TGp_j\right)<0$}
	\EndIf
	and set $v^*=\frac{p_1+\gamma p_j}{\sqrt{1+\gamma^2}}$ where 			
	$\gamma=\frac{-2p_1^TGp_j+\sqrt{\Delta}}{2p_j^TGp_j}$

	~~~and $\Delta=4\left[\left(p_1^TGp_j\right)^2-\left(p_1^TGp_1\right)\left(p_j^TGp_j\right)\right]$
	\State {\bf Remove} $p_k$ from list of vectors $p$ and insert 

	$~~~p^*=\frac{p_j-\gamma p_1}{\sqrt{1+\gamma^2}}$ at the end of the list.
	\State {\bf Add} $\bm{v*}$ to the list $\bm{V$}
	\EndFor
	\State {\bf Add} last element of list $\bm{p}$ to list $\bm{V}$ and 
	\Return $\bm{V}$
\EndFunction
\end{algorithmic}
\end{algorithm}
\end{minipage}
\hfill
\begin{minipage}[c]{0.20 \textwidth}
\end{minipage}
\vspace{10pt}

Algorithm \ref{neuraldecomposition} takes as inputs $Z^*=Z^{k*}_+$ and $Z^*=Z^{k*}_-$ for $k=1,\ldots,p$, which are the solution of the optimization problem (\ref{convextraining}) and thus satisfy ${\rm \bf Trace}\left(Z^*G\right)=0$.
At the end of the algorithm there will be two lists of vectors for each value of $k=1,\ldots,p$. 
One will be the list $\{v^{k,1}_+,\ldots,v^{k,r_k^+}_+\}$ for $Z^*=Z^{k*}_+$ and the other will be the list $\{v^{k,1}_-,\ldots,v^{k,r_k^-}_-\}$ for $Z^*=Z^{k*}_-$ where
\begin{eqnarray}
v^{k,j}_+=\left[
\begin{array}{c}
c^{k,j}_+\\
d^{k,j}_+
\end{array}
\right],
\quad
v^{k,j}_-=\left[
\begin{array}{c}
c^{k,j}_-\\
d^{k,j}_-
\end{array}
\right], \nonumber\\
c^{k,j}_+, c^{k,j}_-\in\RE^n, \quad d^{k,j}_+, d^{k,j}_-\in\RE.
\end{eqnarray}
For each output $k=1,\ldots,p,$ the weights in the first layer of the neural network will be
$(w^{k,1^+},\ldots,w^{k,r_k^+}, w^{k,1^-},\ldots,w^{k,r_k^-})$ where \cite{BartanPilanci2021}
\begin{equation}
w^{k,j^+}=\frac{c^{k,j}_+}{\|c^{k,j}_+\|_2},\quad w^{k,j^-}=\frac{c^{k,j}_-}{\|c^{k,j}_-\|_2}.
\end{equation}
The weights of the second layer of the neural network will be
$(\alpha_1^{k^+},\ldots,\alpha_{r_k^+}^{k^+}, \alpha_1^{k^-},\ldots,\alpha_{r_k^-}^{k^-})$ where \cite{BartanPilanci2021}
\begin{equation}
\alpha_j^{k^+}=\left(d^{k,j}_+\right)^2,\quad \alpha_j^{k^-}=-\left(d^{k,j}_-\right)^2.
\end{equation}
The output $k$ of the neural network for the input $x$ is then given by
\begin{equation}\label{networkoutputk}
\hat y^k(x) = \hat f^k(x) = \sum_{j=1}^{r_k^+}\sigma\left(x^Tw^{k,j^+}\right)\alpha_j^{k^+} +
\sum_{j=1}^{r_k^-}\sigma\left(x^Tw^{k,j^-}\right)\alpha_j^{k^-}
\end{equation}
Renumbering the weights $w^{k,j^-}$ and $\alpha_j^{k^-}$ to run from $j=r_k^++1$ to $j=r_k^++r_k^-$ we can rewrite (\ref{networkoutputk}) as (\ref{networkoutput}) with $M_k=r_k^++r_k^-$.
\vspace{10pt}
\begin{remark}
The convex formulation of the neural network training (\ref{convextraining}) followed by the neural decomposition algorithm \ref{neuraldecomposition} yields a network architecture in which any given hidden layer neuron is only  connected to one output (see figure \ref{neuralnetwork} for the output $\hat y^1$).
This is in contrast to the primal training problem (\ref{nonconvextraining}) which assumed the general case of all hidden layer neurons connected to all outputs.
It is proved in reference \cite{BartanPilanci2021} that the solution of the convex training problem yields the same optimal value for the objective function as the primal training problem if $M\ge M_*$.
\end{remark}
\vspace{10pt}

Instead of describing the neural network by its weights, in this paper we will focus on the description of the neural network as a quadratic form because such a description leads to an analytical expression, namely the one in equation (\ref{networkexpression}) where $\bar x=[x^T~~1]^T$.
The quadratic form (\ref{networkexpression}) is equivalent to the quadratic form in (\ref{convextraining}) and to the expression (\ref{networkoutput}) (see \cite{BartanPilanci2021}).
\begin{equation}\label{networkexpression}
\hat y^k=\hat f^k(x)
=\bar x^T\left[
\begin{array}{ll}
aZ_1^k & \frac{b}{2}Z_2^k\\
\frac{b}{2}\left(Z_2^k\right)^T & cZ_4^k
\end{array}
\right]\bar x=\bar x\bar Z^k\bar x
\end{equation}

The next results introduce standard properties of symmetric matrices.
\vspace{10pt}

\begin{lemma}\label{symmetricdifference}\cite{Algebrabook}
Any real symmetric matrix $Z=Z^T$ can be written as the difference of two real symmetric positive semidefinite matrices, i.e., it can be written as
\begin{equation}\label{psddifference}
Z = Z_+ - Z_-
\end{equation}
where $Z_+\ge 0$ and $Z_-\ge 0$.
\end{lemma}
\vspace{10pt}
\begin{proof}
(page 119 of \cite{Algebrabook})
\end{proof}
For any $\epsilon>0$ one can always write
\begin{equation}
Z = \left(Z + \epsilon I\right) - \epsilon I
\end{equation}
where $I$ is the identity matrix of the same dimension as $Z$.
For sufficiently large $\epsilon$ the (diagonally dominant) matrix $\left(Z +\epsilon I\right)$ is positive semidefinite and $\epsilon I$ is positive definite (therefore semidefinite), which finishes the proof. \done
\vspace{10pt}
\begin{remark} Notice that the result of Lemma \ref{symmetricdifference} is also valid for a scalar since it is a $1\times 1$ matrix. Additionally, any row or colum vector can be written as the difference of two vectors for which all entries are non-negative.
\end{remark}
\vspace{10pt}
\begin{lemma}\cite{Wolkowicz-Styan1980}\label{tracebound}
The maximum eigenvalue of a symmetric matrix $Z=Z^T$ satisfies $\lambda_{max}\left(Z\right)\le {\rm \bf Trace}\left(Z\right)$.
\done
\end{lemma}
\vspace{10pt}
\begin{lemma}\label{expectedvaluequadraticform}\cite{StochasticBook}
Given a symmetric matrix $P$ and a random vector $x_*$ with mean $\mu$ and covariance matrix $\Sigma$,
\begin{equation}\label{expectedvaluewithmeanandcovariance}
E\left[x_*^TPx_*\right]=\mu^TP\mu+{\rm \bf Trace}\left(P\Sigma\right).
\end{equation}
\end{lemma}
\vspace{10pt}
\begin{proof}
See Appendix B of \cite{StochasticBook}.
\end{proof}
\vspace{10pt}

We end this section with three results on the relationship of quadratic neural networks with quadratic forms.
\vspace{10pt}

\begin{theorem}\label{netquadraticequivalence}
Let $\bar x=[x^T~1]^T$ with $x\in\RE^n$, and let $f^k(z)=z^T\bar Z^kz$  with $\bar Z^k=\left(\bar Z^k\right)^T\in\RE^{(n+1)\times(n+1)}$. Given parameters $a\neq 0,~b,~c,$ the quadratic form $f^k(z)$ evaluated at $z=\bar x$ represents an output of a quadratic neural network with activation function (\ref{activationfunction}) if and only if $\bar Z_{n+1,n+1}^k=\frac{c}{a}{\rm \bf Trace}\left(\bar Z_{1:n,1:n}^k\right)$,
where $\bar Z_{n+1,n+1}^k$ is the $(n+1)$-th diagonal element of $\bar Z^k$, and $\bar Z_{1:n,1:n}^k\in\RE^{n\times n}$ is the top left block submatrix of $\bar Z^k$.
\end{theorem}
\vspace{10pt}
\begin{proof}
The proof of the only if statement follows trivially from expression (\ref{networkexpression}).
To prove the if statement we assume that a quadratic form is given.
Since we know $a, b, c,$ we can then compute the values of $Z^{k,1}_+-Z^{k,1}_-, Z^{k,2}_+-Z^{k,2}_-$, and $Z^{k,4}_+-Z^{k,4}_-$ in expression (\ref{networkexpression}).
Additionally, we are assuming that $Z^{k,4}_+-Z^{k,4}_-={\rm \bf Trace}\left[Z^{k,1}_+-Z^{k,1}_-\right]$.
What remains to prove is that there is no additional constraint relating $Z^{k,2}_+-Z^{k,2}_-$ with $Z^{k,4}_+-Z^{k,4}_-$ and $Z^{k,1}_+-Z^{k,1}_-$ for the quadratic form to belong to the feasible set of the optimization (\ref{convextraining}).
We observe that the only additional optimization constraint is the positive semidefiniteness constraints of two matrices.
By the Schur complement the positive semidefiniteness constraints from (\ref{convextraining}) are equivalent to
\begin{eqnarray*}
Z^{k,4}_+={\rm \bf Trace}\left(Z^{k,1}_+\right)\ge 0,\label{traceconstraint}\\
\left[1-{\rm \bf Trace}\left(Z^{k,1}_+\right){\rm \bf Trace}^\dagger\left(Z^{k,1}_+\right)\right]
\left(Z^{k,2}_+\right)^T=0,\label{equalityconstraint}\\
Z^{k,1}_+-Z^{k,2}_+{\rm \bf Trace}^\dagger\left(Z^{k,1}_+\right)\left(Z^{k,2}_+\right)^T\ge 0,
\label{constraintonZ2}
\end{eqnarray*}
where, defining $t_k^+={\rm \bf Trace}\left(Z^{k,1}_+\right)$,
\begin{equation*}\label{tracedagger}
{\rm \bf Trace}^\dagger\left(Z^{k,1}_+\right)=\left\{
\begin{array}{rcl}
0,&{\rm if}&~t_k^+=0,\\
{\rm \bf Trace}^{-1}\left(Z^{k,1}_+\right),&{\rm if}&~t_k^+\neq 0,
\end{array}
\right.
\end{equation*}
is the Moore-Penrose pseudo-inverse of ${\rm \bf Trace}\left(Z^{k,1}_+\right)$, with
the same conditions applying for the case of $Z^{k,1}_-,~Z^{k,2}_-$.
Simple algebraic manipulations then lead to
\begin{eqnarray*}
Z^{k,2}_+\left(Z^{k,2}_+\right)^T &\le& Z^{k,1}_+\left[{\rm \bf Trace}\left(Z^{k,1}_+\right)\right],\label{boundZ2plus}\\
Z^{k,2}_-\left(Z^{k,2}_-\right)^T &\le& Z^{k,1}_-\left[{\rm \bf Trace}\left(Z^{k,1}_-\right)\right].\label{boundZ2minus}\\
\end{eqnarray*}
Applying the trace operator to these constraints leads to
\begin{eqnarray}\label{normbound}
\|Z^{k,2}_+-Z^{k,2}_-\|^2\le\left(\|Z^{k,2}_+\|+\|Z^{k,2}_-\|\right)^2\nonumber\\
\le\left[{\rm \bf Trace}(Z^{k,1}_+)+{\rm \bf Trace}(Z^{k,1}_-)\right]^2=\left[Z^{k,4}_++Z^{k,4}_-\right]^2.
\end{eqnarray}
We thus observe that $\|Z^{k,2}_+-Z^{k,2}_-\|$ is constrained by $Z^{k,4}_++Z^{k,4}_-$.
Note however that although the value of $Z^{k,4}_+-Z^{k,4}_-$ is constrained to be fixed given a quadratic form written as (\ref{networkexpression}) and parameters $a,b,c$, the value of $Z^{k,4}_++Z^{k,4}_-$ is arbitrary.
Therefore, from (\ref{normbound}) we see that there is no constraint relating $Z^{k,1}_+-Z^{k,1}_-$ or $Z^{k,4}_+-Z^{k,4}_-$ with $Z^{k,2}_+-Z^{k,2}_-$, which finishes the proof.
\end{proof}
\vspace{10pt}

We can conclude from the results of Theorem \ref{regularizationresult} that if one wants all the entries in the parameter matrices $Z_k^+$ and $Z_k^-$ to be bounded it suffices to penalize large values of the traces of $Z_{k,1}^+$ and $Z_{k,1}^-$.
This can be done either by adding a constraint or by a process called regularization, which adds a penalty in the cost for high values of these parameters.
Note that the regularization is included in the convex program (\ref{convextraining}) but not the constraint.
The regularization corresponds to the term on the objective function that multiplies $\beta\ge 0$.
This term is equal to the sum of the traces of $Z_k^+$ and $Z_k^-$, which are non-negative because the matrices are positive semidefinite. 
According to Lemma \ref{tracebound}, the maximum eigenvalue of a symmetric matrix is bounded by its trace.
Therefore, the regularization term in the convex program (\ref{convextraining}) penalizes an upper bound on the sum of the maximum eigenvalues of the matrices $Z_{k,1}^+$ and $Z_{k,1}^-$, which corresponds to the sum of the norm-$2$ of the matrices.

The analysis of the constraints on the network parameters for the case of a single input and single output (SISO) neural network is now stated as a corollary of Theorem \ref{netquadraticequivalence}.
\vspace{10pt}
\begin{corollary}\label{regularizationresult}
For a quadratic neural network with a single input and a single output the parameters of the network are constrained by
\begin{eqnarray}\label{plusconstraints}
Z^1_+ &\ge 0,\nonumber\\
|Z^2_+| &\le& Z^1_+,\nonumber\\
 Z^4_+ &=& Z^1_+,
\end{eqnarray}
and
\begin{eqnarray}\label{minusconstraints}
Z^1_- &\ge 0,\nonumber\\
|Z^2_-| &\le& Z^1_-,\nonumber\\
 Z^4_- &=& Z^1_-.
\end{eqnarray}
Moreover, the network output cannot be an affine polynomial.
\end{corollary}
\vspace{10pt}
\begin{proof}
For a single input single output network the matrix in equation (\ref{networkexpression}) is $2\times2$. From (\ref{convextraining}) (ommiting $k=1$) we get
\begin{equation}\label{networksingleoutput}
Z^4_+=Z^1_+,\quad Z^4_-=Z^1_-,
\end{equation}
because the trace of a scalar is the scalar itself.
This in turn implies that the output of the neural network is $\hat y=aZ_1^1x^2+bZ_2^1x+\frac{c}{a}Z^1_1$ where $Z^1_1=Z^1_+-Z^1_-$, $Z^1_2=Z^2_+-Z^2_-$.
Therefore, when $c\neq 0$ the network output will never be an affine polynomial because the output is linear if $Z^1_1=0$ and it is quadratic if $Z^1_1\neq0$.
If $c=0$, the affine term is zero and therefore the network output can also not be an affine polynomial.
From (\ref{convextraining}) and (\ref{networksingleoutput}), the positive semidefinite constraints on $Z_+$ and $Z_-$ lead to (\ref{plusconstraints}) and (\ref{minusconstraints}), respectively, which finishes the proof. 
\end{proof}
\vspace{10pt}
\begin{example}
The quadratic form
\begin{equation*}
f(z)=z^T\bar Zz,\quad
\bar Z = \left[
\begin{array}{cc}
0 & 0\\
0 & 1
\end{array}
\right],
\end{equation*}
cannot represent a SISO quadratic neural network because the second diagonal entry is not equal to a multiple of the first diagonal entry, which is zero, thus violating the condition of Theorem \ref{netquadraticequivalence}. \done
\end{example}
\vspace{10pt}
\begin{theorem}\label{Lipschitz}
Each output of a quadratic neural network satisfies the following Lipschitz inequality
\begin{equation}\label{Lipschitzcondition}
|\hat f^k(x_1)-\hat f^k(x_2)|\le L_n\left(\bar x_1,\bar x_2,\bar Z^k\right)\|\bar x_1-\bar x_2\|_2,~\forall x_1,~x_2\in\RE^{n},
\end{equation}
where
\begin{eqnarray*}
L_n\left(\bar x_1,\bar x_2,\bar Z^k\right)=\sqrt{n+1}~\left|\lambda_{max}\left(\bar Z^k\right)\right|\left(\|\bar x_1\|_{\infty}+\|\bar x_2\|_{\infty}\right)
\end{eqnarray*}
\end{theorem}
\vspace{10pt}
\begin{proof}
Using the expression (\ref{networkexpression}) for the output of a quadratic neural network one can write
\begin{eqnarray*}
|\hat f^k(x_1)-\hat f^k(x_2)| =
 |\left(\bar x_1 + \bar x_2\right)^T\bar Z^k\left(\bar x_1 - \bar x_2\right)|.
\end{eqnarray*}
Using the Cauchy-Schwartz and the triangular inequalities yields
\begin{eqnarray}
|\hat f^k(x_1)-\hat f^k(x_2)| \le\left(\|\bar x_1\|_2 + \|\bar x_2\|_2\right)\|\bar Z^k\|_2\|\bar x_1 - \bar x_2\|_2.
\end{eqnarray}
The result then follows since $\|\bar x\|_2\le\sqrt{n+1}~\|\bar x\|_{\infty}$, for all $x\in\RE^{n}$ and $\|\bar Z^k\|_2=\left|\lambda_{max}\left(\bar Z^k\right)\right|$.
\end{proof}
\vspace{10pt}
\begin{remark}
Notice that if the input $x$ of the neural network is normalized so that its infinity norm is bounded by a constant, then the inequality (\ref{Lipschitzcondition}) provides a Lipschitz continuity constant for each output of the neural network.
This means that for inputs that are ''close enough'' to a given input the outputs will also be ''close enough''.
Therefore, the network is resillient and robust to small perturbations of the input.
\end{remark}
\vspace{10pt}
\begin{remark}
From the constraints of the optimization (\ref{convextraining}) we conclude that $T^k={\rm \bf Trace}\left(\bar Z^k\right)=(a+c)\left(Z^{k,4}_+ - Z^{k,4}_-\right)$.
From the result of Lemma \ref{tracebound} one can then conclude that $\left|\lambda_{max}\left(\bar Z^k\right)\right|\le|T^k|\le|a+c|\left(Z^{k,4}_+ + Z^{k,4}_-\right)$ when $\lambda_{max}\left(\bar Z^k\right)\ge 0$.
In that case, the term multiplying the regularization parameter $\beta$ in the objective function of problem (\ref{convextraining}) is weighting an upper bound on the sum of the Lipschitz constants for all network outputs. Therefore, the optimization problem (\ref{convextraining}) is minimizing an objective that is a trade-off between the accuracy of the network on the training set and the sum of upper bounds on the Lipschitz constants for all outputs. If the parameter $\beta$ is small, one can expect high accuracy and when it is large one can expect that the Lipschitz constants will be small. This is a similar trade-off as the one proposed in \cite{Allgoweretall2022}.
\end{remark}

\section{Regression, Classification, and System Identification}\label{regressionclassification}
This section addresses the applications of quadratic neural networks to regression, classification, and system identification.

\subsection{Regression}
To perform regression on pairs of data vectors $\{(x_i,y_i)\}_{i=1}^N$ we define the matrices
\begin{equation}\label{regressionmatrices}
X = \left[
\begin{array}{c}
x^T_1\\
\vdots\\
x^T_N
\end{array}
\right],\quad
Y=\left[
\begin{array}{c}
y^T_1 \\
\vdots\\
y^T_N
\end{array}
\right].
\end{equation}
Each row of the matrix $X$ is an input sample and each row of the matrix $Y$ is an output label of the training set of the quadratic neural network.
These two matrices together with the constant parameters $a, b, c,$ and $\beta$ are the input to the training of the neural network. If one wants to add offset weights to the hidden layer then instead of using $X$ as the input one should use $\mathcal X=[X~~1_{N\times 1}]$, where $1_{N\times 1}$ is a column vector of ones. 
The output of the training is the optimal network parameters $\bar Z^k, k=1,\ldots,p$, which can be transformed into neuron weights using the neural decomposition algorithm \ref{neuraldecomposition} \cite{BartanPilanci2021}.

\subsection{Classification}
An important application for quadratic neural networks is the classification of a given input into several possible classes. A typical example is the recognition of handwritten digits between $0$ and $9$. To do this, the standard approach is to have as many outputs as there are classes (digits in this case) and to have the desired output label be either $0$ or $1$ for each output, depending on which digit is provided as the input of the network. For example, if the input digit is $0$ then the first output label of the network should be $1$ and all other nine output labels should be $0$. After the network is trained, an extra block is added to the output of the network to perform the operation $\arg\max(\cdot)$. In other words, this extra block determines which output has the maximum value. The  $\arg\max(\cdot)$ block effectively adds a nonlinearity to the output layer. Handwritten digit classification examples are shown in section \ref{examples}.

\subsection{System Identification}\label{modeling}
Quadratic neural networks can be used to obtain input-output and state space models of a discrete-time dynamical system.
It is assumed that a collection of input output data pairs $\{u(k),y(k)\}_{k=1}^N$ are measured with $N\gg 1$.
Based on the data one can identify the parameters of an autoregressive model of the form
\begin{equation}\label{autoregressive}
y(t+1) = f(y(t-n+1),\ldots,y(t),u(t))
\end{equation}
by training a quadratic neural network, where $y\in\RE^p,~u\in\RE^m$, and $n\ge 1$. The value of $n-1$ gives the number of delays considered in the output.
If $n=1$ then the model only considers the most recent sample $y(t)$. When $n>1$ the model will also consider past samples of $y$.
To have an initial idea of how many samples should be considered in the past to identify the autoregressive model one can look at the autocorrelation of the measured output and see how it decays as a function of the shift (delay) in the output.
The delay at which the first significant decay happens should be the maximum delay used in the identification.
We define the training matrices as
\begin{equation}\label{systemidmatrices}
X = \left[
\begin{array}{cccc}
u^T(n) & y^T(1) & \ldots & y^T(n)\\
\vdots &\vdots &\vdots &\vdots\\
u^T(N-1) & y^T(N-n) & \ldots & y^T(N-1)
\end{array}
\right],\quad
Y=\left[
\begin{array}{c}
y^T(n+1) \\
\vdots\\
y^T(N)
\end{array}
\right],
\end{equation}
where $X\in\RE^{(N-n)\times(m+pn)}$ and $Y\in\RE^{(N-n)\times p}$.
As before, each row of the matrix $X$ is a neural network input sample and each row of the matrix $Y$ is an output label of the training set of the quadratic neural network.
It is assumed that $N\ge n+0.5(pn+m+1)(pn+m+2)$ so that the parameter estimation problem is overdetermined.
It is also assumed that the collected data is rich enough in terms of persistent excitation \cite{Ljungbook}, which is a typical assumption in system identification.
After training the network, the input-output model is written as in equation (6) changing $\bar x(t)$ to $\bar y_u(t)=[u^T(t)~~y^T(t-n+1)~~\ldots~~y^T(t)~~1]^T$.
Defining state variables as $x_1(t)=y(t-n+1),~\ldots,~x_n(t)=y(t),$ a state vector $x(t)=[x_1^T(t),\ldots,x_n^T(t)]^T$, and noting that $\bar y_u(t)=[u^T(t)~~x^T(t)~~1]^T=[u^T(t)~~\bar x^T(t)]^T$, the state space model is written as
\begin{equation}\label{statespacemodel}
x(t+1)=Ax(t)+g(x(t),u(t))=\left[
\begin{array}{ll}
0_{p(n-1)\times p} & I_{p(n-1)\times p(n-1)}\\
0_{p\times p} & 0_{p\times p(n-1)}
\end{array}
\right]x(t)+
\left[
\begin{array}{c}
0_{p(n-1)\times 1}\\
\mathcal Z(x(t),u(t))
\end{array}
\right],
\end{equation}
where
\begin{equation}\label{mathcalZ}
\mathcal Z(x(t),u(t))=\left[
\begin{array}{c}
y^1(t+1)\\
\vdots\\
y^p(t+1)
\end{array}
\right]=
\left[
\begin{array}{c}
\bar y_u^T(t)\bar Z^1\\
\vdots\\
\bar y_u^T(t)\bar Z^p
\end{array}
\right]\bar y_u(t),\quad
\bar Z^i=\left[
\begin{array}{ccc}
\bar Z_{uu}^i & \bar Z_{ux}^i & \bar Z_{u}^i\\
\left(\bar Z_{ux}^i\right)^T & \bar Z_{xx}^i & \bar Z_{x}^i\\
\left(\bar Z_{u}^i\right)^T & \left(\bar Z_{x}^i\right)^T & \bar Z_{nn}^i
\end{array}
\right],~i=1,\ldots,p.
\end{equation}
Expanding the quadratic forms in $\mathcal Z(x(t),u(t))$, the system (\ref{statespacemodel})--(\ref{mathcalZ}) can be rewritten as
\begin{equation}\label{statespacemodel2}
\bar x(t+1) = \bar A(x(t))\bar x(t) + \bar B(x(t)) u(t) + E(u(t))u(t),
\end{equation}
where $x\in\RE^{n_x}$ with $n_x=pn$, $u\in\RE^m$, $\bar A(x(t))=\mathcal A+F(x(t))$ and
\begin{eqnarray}\label{modelmatrices}
\mathcal A=\left[
\begin{array}{cl}
A & 0_{pn\times 1}\\
0_{1\times pn} &  1
\end{array}
\right],
A=\left[
\begin{array}{ll}
0_{p(n-1)\times p} & I_{p(n-1)\times p(n-1)}\\
0_{p\times p} & 0_{p\times p(n-1)}
\end{array}
\right],
\bar Z^i_{\bar x\bar x}=\left[
\begin{array}{cc}
\bar Z_{xx}^i & \bar Z_{x}^i\\
\left(\bar Z_{x}^i\right)^T & \bar Z_{nn}^i
\end{array}
\right], \bar Z^i_{u\bar x}=\left[\bar Z_{ux}^i~~\bar Z_{u}^i\right],\nonumber\\ 
F(x(t))=\left[
\begin{array}{c}
0_{p(n-1)\times (pn+1)}\\
\bar x^T(t)\bar Z^1_{\bar x\bar x}\\
\vdots\\
\bar x^T(t)\bar Z^p_{\bar x\bar x}\\
0_{1\times (pn+1)}
\end{array}
\right],
E(u(t))=\left[
\begin{array}{c}
0_{p(n-1)\times m}\\
u^T(t) \bar Z^1_{uu}\\
\vdots\\
u^T(t) \bar Z^p_{uu}\\
0_{1\times m}
\end{array}
\right],
\bar B(x(t))=\left[
\begin{array}{c}
B(x(t))\\
0_{1\times m}
\end{array}
\right]=
2\left[
\begin{array}{c}
0_{p(n-1)\times m}\\
\bar x^T(t) \left[\bar Z^1_{u\bar x}\right]^T\\
\vdots\\
\bar x^T(t)\left[\bar Z^p_{u\bar x}\right]^T\\
0_{1\times m}
\end{array}
\right].
\end{eqnarray}

\section{Lyapunov Control}\label{controlsystems}
This section is devoted to controller synthesis.
The results about to be derived are applicable to a system model of the form (\ref{statespacemodel2}), which includes but is not limited to quadratic neural networks under the formulation (\ref{statespacemodel})--(\ref{mathcalZ}).
Before describing the design methodology, an important observation to be made is that the model (\ref{statespacemodel2}) is nonlinear. Among other phenomena, it may have multiple steady state solutions for the control input, as shown in example \ref{firstordersystem} of section \ref{examples}.
For a system or a neural network modelled by equation (\ref{statespacemodel2}) we propose to design a controller
\begin{equation}\label{control}
u(t)=K(x(t))(x(t)-x_\star)+u_\star=K_{x_t}(x(t)-x_\star)+u_\star=\bar K_{x_t}\bar x(t)
\end{equation}
to stabilize the closed-loop system to a desired state $x_*$, where $u_\star$ is the input steady state value when $x(t)=x_\star$ and
\begin{equation}\label{Kbar}
\bar K_{x_t}=\left[K_{x_t}~~u_*-K_{x_t}x_*\right].
\end{equation}
Replacing the control input (\ref{control}) in (\ref{statespacemodel2}) yields
\begin{equation}\label{ClosedLoop}
\bar x(t+1) = A_{cl}(x(t),\bar K_{x_t})\bar x(t),
\end{equation}
where
\begin{equation}\label{Acl}
A_{cl}(x(t),\bar K_{x_t}) = \bar A(x(t))+\bar B(x(t))\bar K_{x_t}+E(\bar K_{x_t}\bar x(t))\bar K_{x_t}.
\end{equation}
We first design $u_*$ based on the desired steady state response of the system and then design $K_{x_t}$ using a control Lyapunov function that guarantees closed-loop stability.

\subsection{Steady State Input}
When a system model is in the form (\ref{statespacemodel2}) and the desired setpoint for the steady state $x_*$ is given, one must solve the equation
\begin{equation}\label{steadystateequationstar}
\bar x_* = \bar A(x_*)\bar x_* + \bar B(x_*) u_* + E(u_*)u_*
\end{equation}
to determine the steady state value of the input $u_*$.
This is illustrated in example \ref{firstordersystem} in section \ref{examples} where the steady state input $u_*$ has two solutions.
Note that if the steady state value of $u_*$ is different from zero it implies that there must be a constant control input once the system achieves the desired steady state setpoint $x_*$.
Physical systems typically have at least one equilibrium point for which no control action is needed to keep the system in equilibrium.
For example, a mass-spring system has a zero input force equilibrium point for a deflection of the spring equal to zero.
Based on this observation we will make the assumption that the coordinates $x(t)$ were chosen in such a way that $x_*=0$ is a zero input ($u_*=0$) equilibrium state.
\vspace{10pt}

\noindent{\bf Assumption 1}: The input $u_*=0$ is a solution of (\ref{steadystateequationstar}) when $x_*=0$.
\vspace{10pt}

For the case where the system model is obtained from an input-output system identification and is in the form (\ref{statespacemodel}) with $x(t)=[y^T(t-n+1)~~\ldots~~y^T(t)]^T$, we assume that a desired setpoint $y_*$ for the output is given.
The desired state setpoint will then be $x_*=\Gamma y_\star$, where $\Gamma\in\RE^{np\times p}$ is written as
\begin{equation}\label{steadystatematrix}
\Gamma = \left[
\begin{array}{c}
I_p\\
\vdots\\
I_p
\end{array}
\right],
\end{equation}
where $I_p$ is the identity matrix of order $p$.
The steady state value of each output is $y^i_{ss}=y_*^i$, where $y_*^i$ is the $i$-th coordinate of the desired steady state  output vector $y_*$.
Therefore, from (\ref{statespacemodel})--(\ref{mathcalZ}) one can write
\begin{equation}\label{steadystateequation}
y_*^i=\bar y_{u*}^T\bar Z^i\bar y_{u*},~i=1,\ldots,p,
\end{equation}
where 
\begin{equation}\label{steadybarstate}
\bar y_{u*}(t)=\left[
\begin{array}{c}
u_*\\
\Gamma y_*\\
1
\end{array}
\right].
\end{equation}
Equations (\ref{steadystateequation})--(\ref{steadybarstate}) must be solved for $u_*$ given a desired setpoint $y_*$ in order to find the steady state values of the control input.
Notice from (\ref{steadystateequation})--(\ref{steadybarstate}) and (\ref{mathcalZ}) that if $\bar Z_{nn}^i=0$ (no constant offset terms) then $u_*=0$ will be a solution of (\ref{steadystateequation}) when $y_*=0$ and assumption 1 will be satisfied.

\subsection{Lyapunov Controller Synthesis}
After computing a solution of (\ref{steadystateequation})--(\ref{steadybarstate}), when one exists, replacing the input (\ref{control}) in equation (\ref{statespacemodel}) yields
\begin{equation}\label{closedloopequation}
y^i(t)=\bar x^T(t)\bar Z_{cl}^i(\bar K_{x_t})\bar x(t),~i=1,\ldots,p,
\end{equation}
where
\begin{equation}\label{closedloopbarZ}
\bar Z_{cl}^i(\bar K_{x_t})=\left[
\begin{array}{cc}
\bar Z_{x_{cl}x_{cl}}^i & \bar Z_{x_{cl}}^i \\
\left(\bar Z_{x_{cl}}^i\right)^T & \bar Z_{n_{cl}}^i
\end{array}
\right],~i=1,\ldots,p,
\end{equation}
with
\begin{eqnarray}
Z_{x_{cl}x_{cl}}^i &=& \bar Z_{xx}^i + K_{x_t}^T\bar Z_{ux}^i + \left[\bar Z_{ux}^i\right]^T K_{x_t} + K_{x_t}^T\bar Z_{uu}^iK_{x_t},\\
\bar Z_{x_{cl}}^i &=& \bar Z_{x}^i + \left(\bar Z_{uu}^iK_{x_t}+\bar Z_{ux}^i\right)^T\left(u_*-K_{x_t}x_*\right)+K^T_{x_t}\bar Z_u^i,\\
\bar Z_{n_{cl}}^i &=&\bar Z_{nn}^i + 2\left(u_*-K_{x_t}x_*\right)^TZ_u^i + \left(u_*-K_{x_t}x_*\right)^T\bar Z_{uu}^i\left(u_*-K_{x_t}x_*\right).
\end{eqnarray}
Therefore, the ouput is written as
\begin{equation}\label{outputclosedloop}
y(t)=\left[
\begin{array}{c}
\bar x^T(t)\bar Z_{cl}^1(\bar K_{x_t})\\
\vdots\\
\bar x^T(t)\bar Z_{cl}^p(\bar K_{x_t})
\end{array}
\right]\bar x(t).
\end{equation}
From (\ref{statespacemodel}), (\ref{mathcalZ}), and (\ref{outputclosedloop})
the closed-loop state space model can be rewritten as in equation (\ref{ClosedLoop}) 
where
\begin{equation}\label{M}
A_{cl}(x(t),\bar K_{x_t})=\left[
\begin{array}{c}
\bar I\\
\bar x^T(t)Z_{cl}^1(\bar K_{x_t})\\
\vdots\\
\bar x^T(t)Z_{cl}^p(\bar K_{x_t})\\
\bar 0^T
\end{array}
\right],\quad \bar I=\left[0_{p(n-1)\times p}~~I_{p(n-1)\times p(n-1)}~~0_{p(n-1)\times 1}\right],\quad
\bar 0^T=[0_{1\times pn}~~1].
\end{equation}
Equation (\ref{M}) can be rewritten in the form (\ref{Acl}).
Making the substitution $\bar K_{x_t}=0$ in (\ref{M}) yields the open-loop model.

Stability of the closed-loop system for a given controller can be proved by finding a Lyapunov function.
The next Theorem provides a Lyapunov-based design strategy that yields a provably stabilizing controller.
\vspace{10pt}

\begin{theorem} Given a desired setpoint $x_*\in\mathcal X\subseteq\RE^{n_x}$, if the inequality
\begin{equation}\label{newdecreasingcondition}
A_{cl}^T(x(t),\bar K_{x_t}) \bar PA_{cl}(x(t),\bar K_{x_t})-\bar P \le 0,~~\forall~x(t)\neq  x_*, x(t)\in\mathcal X\subseteq\RE^{n_x},
\end{equation}
is satisfied where $A_{cl}$ is given by (\ref{Acl}) [or by (\ref{M})] and $\bar P$ is defined as
\begin{equation}\label{barP}
\bar P=\left[
\begin{array}{rr}
P \quad& -Px_*\\
-x_*^TP \quad & x_*^TPx_*
\end{array}
\right]
\end{equation}
where $P>0$, then the controller $u(t)=\bar K_{x_t}\bar x(t)$ renders the closed-loop system (\ref{ClosedLoop}) stable in the sense of Lyapunov inside the largest invariant set of the Lyapunov function (\ref{Lyapunovfunction}) fully contained in $\mathcal X$.
If the inequality (\ref{newdecreasingcondition}) is strict, the closed-loop system is asymptotically stable inside the same invariant set.
If $\mathcal X=\RE^{n_x}$, the stability is global.
\end{theorem}
\vspace{10pt}
\begin{proof}
If $x_*$ is the desired setpoint in steady state we define the candidate quadratic control Lyapunov function
\begin{equation}\label{Lyapunovfunction}
V(x(t))=\left(x(t)-x_*\right)^TP\left(x(t)-x_*\right)=\bar x^T(t)\bar P\bar x(t),
\end{equation}
where $\bar P$ is defined in (\ref{barP}).
For Lyapunov stability one must satisfy the conditions
\begin{eqnarray}
V(x_*) = 0,&& \label{equilibrium}\\
V(x) > 0,&&~\forall~x\neq x_*, x\in\mathcal X\label{positivedefinite}\\
\Delta V = V(x(t+1)) - V(x(t)) \le 0, &&~\forall~x(t)\neq x_*, x(t)\in\mathcal X.\label{decreasing}
\end{eqnarray}
Notice from the definition of the Lyapunov function (\ref{Lyapunovfunction}) that the condition (\ref{equilibrium}) is guaranteed to be satisfied.
Furthermore, the condition (\ref{positivedefinite}) is also satisfied because $P>0$.
Computing $V(x(t+1))$ using (\ref{ClosedLoop}) yields
\begin{equation}\label{Lyapunovnexttime}
V(x(t+1))=\bar x^T(t)A_{cl}^T(x(t),\bar K_{x_t}) \bar PA_{cl}(x(t),\bar K_{x_t})  \bar x(t).
\end{equation}
Using (\ref{Lyapunovnexttime}) condition (\ref{decreasing}) is implied by (\ref{newdecreasingcondition}).
From standard Lyapunov theory the system is then stable inside the largest invariant set of the Lyapunov function (\ref{Lyapunovfunction}) fully contained in $\mathcal X$.
The proof of asymptotical stability follows by noting that the Lyapunov function is decreasing when the strict inequality is satisfied.
Global stability follows from the fact that the Lyapunov function (\ref{Lyapunovfunction}) is radially unbounded.
\end{proof}
\vspace{10pt}

\begin{remark}\label{nobarcase}
Notice that under assumption 1 the last column of $A_{cl}$ is a vector with all entries zero except the last, which is equal to one. Moreover, if $x_*=0$ then $\bar P$ defined in (\ref{barP}) has the last row and the last column equal to the zero vector.
Therefore ,the condition (\ref{newdecreasingcondition}) is equivalent  in this case to the same inequality being satisfied by the upper left matrix block obtained after removing the last row and the last column. Example \ref{simplerexample} in section \ref{examples} will highlight this point.
\end{remark}
\vspace{10pt}

Note that, according to Theorem \ref{netquadraticequivalence}, for a known vector $x_*$ the Lyapunov function (\ref{Lyapunovfunction}) is a quadratic neural network with activation parameters $a\neq 0, b, c,$ if and only if $x_*^TPx_*=\frac{c}{a}{\rm \bf Trace}(P)$. Theorem \ref{Lyapunovonaverage} provides a result for the case of a random vector $x_*$.
\vspace{10pt}

\begin{theorem}\label{Lyapunovonaverage}
Consider the quadratic Lyapunov function (\ref{Lyapunovfunction}) with $P>0$.
Given scalars $c$ and $a\neq 0$, if the desired steady state $x_*$ is a random vector with zero mean and covariance matrix $\Sigma=\frac{c}{a}I$, where $I$ is the identity, then
\begin{equation}\label{expectedvalue}
E\left[x_*^TPx_*\right]=\frac{c}{a}{\rm \bf Trace}\left(P\right).
\end{equation}
In other words, the matrix $\bar P$ in (\ref{barP}) can represent ''on average'' a quadratic neural network with activation function parameters $a\neq 0, b, c,$ if and only if the desired steady state vector $x_*$ is a random vector with zero mean and covariance matrix $\Sigma=\frac{c}{a}I$.
\end{theorem}
\vspace{10pt}
\begin{proof}
It follows from Theorem \ref{netquadraticequivalence} and Lemma \ref{expectedvaluequadraticform} with $\mu=0$ and $\Sigma=\frac{c}{a}I$.
\end{proof}
\vspace{10pt}

Unfortunately, condition (\ref{newdecreasingcondition}) is not convex for a couple of reasons.
First of all, it is not linear in the unknown parameters.
Additionally, unless $A_{cl}$ has polynomial entries in $x(t)$, a sum of squares \cite{Prajna05} argument cannot be used to provide a convex relaxation, as it will be done later in section \ref{convexformulation}.
As such, it is in general difficult to solve inequality  (\ref{newdecreasingcondition}), except in simple problems as illustrated in example \ref{simplerexample} of section \ref{examples}.
\vspace{10pt}

To be able to cast controller synthesis as a convex optimization program in section \ref{convexformulation} we will need the following assumption.
\vspace{10pt}

\noindent {\bf Assumption 2}: $E(u(t))=0$ for all $u(t)\in\RE^m$.
\vspace{10pt}

Under assumptions 1 and 2 the model (\ref{statespacemodel2}) can be rewritten in the form
\begin{equation}\label{statespacemodelinx}
x(t+1) = A(x(t)) x(t) + B(x(t)) u(t),
\end{equation}
for appropriate matrices $A\in\RE^{n_x\times n_x}, B\in\RE^{nx\times m}$.
The following result can then be stated and proved.
\vspace{10pt} 

\begin{theorem}\label{convexsolution}
Assume that a system model is given in the form (\ref{statespacemodelinx})
where $x(t)\in\mathcal X\subseteq\RE^{n_x}$.
If the desired steady state setpoint is $x_*=0$ and if for a given $\epsilon\in(0,1)$ there are $P=P^T$ and $L(x(t))$ satisfying
\begin{equation}\label{convexinequality}
\left[
\begin{array}{cc}
(1-\epsilon)P-\epsilon I & PA^T(x(t)) + L(x(t))^TB^T(x(t))\\
A(x(t)) P+B(x(t))L(x(t)) & P
\end{array}
\right] \ge 0,~\forall~x(t)\in\mathcal X,
\end{equation}
then the closed-loop system is asymptotically stable in the largest invariant set of the Lyapunov function $V(x)=x^TP^{-1}x$ fully contained inside $\mathcal X$.
If $\mathcal X=\RE^{n_x}$ then the stability is global.
The control input is $u(t)=L(x(t))P^{-1}x(t)$.
\end{theorem}
\vspace{10pt}
\begin{proof}
Replacing $u(t)=K(x(t))x(t)$ in (\ref{statespacemodelinx}) yields the closed-loop system 
\begin{equation}\label{Aclaffineinuwithx}
x(t+1)=A_{cl}(x(t),K(x(t))x(t),\quad A_{cl}(x(t), K(x(t)))=A(x(t))+B(x(t))K(x(t)).
\end{equation}
Consider a candidate Lyapunov function of the form $V(x)=x^TP^{-1}x$.
Note that if the condition (\ref{convexinequality}) is satisfied then $P>0$ and therefore it is invertible.
It is clear that $V(0)=0$ and $V(x)>0$ for $x\neq 0$.
Additionally, if (\ref{convexinequality}) is satisfied then applying the Schur complement and using (\ref{Aclaffineinuwithx}) it implies that
\begin{equation}\label{Lyapunovnexttimeconvex2}
PA_{cl}^T(x(t),K(x(t))) P^{-1}A_{cl}(x(t),K(x(t)))P - (1-\epsilon)P\le-\epsilon I<0~\forall~x\in\mathcal X,
\end{equation}
using the substitution $K=L(x(t))P^{-1}$.
Multiplying on the left by $x^T(t)P^{-1}$ and on the right by $P^{-1}x(t)$ yields $\Delta V<-\epsilon x^TP^{-1}x,~\forall x\in\mathcal X$, which guarantees asymptotic stability in the largest level set of $V(x)$ fully contained in $\mathcal X$.
Global stability follows from the fact that the Lyapunov function is radially unbounded.
\end{proof}
\vspace{10pt}
\begin{remark}
The variables in the inequality (\ref{convexinequality}) are $P$ and $L(x)$. Since one must invert $P$ to compute the control input $u(t)=L(x(t))P^{-1}x(t)$ one way to make sure that the numerical inversion is well conditioned is to find the matrix $X$ that satisfies (\ref{convexinequality}) with the minimum condition number. This can be done by adding the constraint
\begin{equation}\label{conditionumber}
P\le\eta I,
\end{equation}
and then minimizing $\eta$.
\end{remark}

\subsection{Convex Formulation of Lyapunov Controller Synthesis}\label{convexformulation}
To formulate controller synthesis as a convex optimization problem we will need the framework of sum of squares polynomials.
For $\textbf{x}\in\RE^n$, a multivariate polynomial $p(\textbf{x})$ is a sum of squares (SOS) if there exist some polynomials $f_i(\textbf{x}),~i=1,\ldots,M$, such that \cite{Prajna05}
\begin{equation}
  p(\textbf{x}) = \sum_{i=1}^M f_i^2(\textbf{x}).
\end{equation}
A polynomial $p(\textbf{x})$ of degree $2d$ is a sum of squares if and only if there exists a positive semidefinite matrix $Q$ and a vector $Z(\textbf{x})$ containing monomials in $\textbf{x}$ of degree less than $d$ such that \cite{Prajna05}
\begin{equation}
  p(\textbf{x}) = W(\textbf{x})^TQW(\textbf{x}).
\end{equation}
It should be noted that if $p(\textbf{x})$ is a sum of squares then $p(\textbf{x})\geq0$, but the converse is generally not true. 
For a convex formulation of controller synthesis the inequality (\ref{convexinequality}) will be relaxed into a sum of squares and the control gain matrix $K(x(t))$ will be constrained to have polynomial entries.
The following corollary of Theorem \ref{convexsolution} can then be stated.
\vspace{10pt}

\begin{corollary}\label{convexsynthesis}
Assume that a system model is given in the form (\ref{statespacemodelinx}) where $A(x(t))$ and $B(x(t))$ have polynomial entries and $x(t)\in\mathcal X\subseteq\RE^{n_x}$.
If the desired steady state setpoint is $x_*=0$ and if for a given $0\le\epsilon< 1$ there are $P>0$ and $L(x(t))$ with polynomial entries satisfying
\begin{equation}\label{convexinequalitysos}
\left[
\begin{array}{cc}
(1-\epsilon)P-\epsilon I & PA(x(t)) + L^T(x(t))B^T(x(t))\\
A^T(x(t)) P+B(x(t))L(x(t)) & P
\end{array}
\right] {\rm is~SOS},~\forall~x(t)\in\mathcal X,
\end{equation}
then the closed-loop system is stable in the largest invariant set of the Lyapunov function $V(x)=x^TP^{-1}x$ fully contained inside $\mathcal X$.
If the condition (\ref{convexinequalitysos}) is satisfied for $\epsilon\in(0,1)$, then the closed-loop system is asymptotically stable inside the same invariant set.
If $\mathcal X=\RE^n$ the stability is global.
The control input is given by $u(t)=L(x(t))P^{-1}x(t)$.
\end{corollary}
\vspace{10pt}
\begin{proof}
The proof follows the argument of the proof of Theorem \ref{convexsolution} upon observing that satisfaction of the condition (\ref{convexinequalitysos}) implies that the inequality (\ref{convexinequality}) is satisfied.
\end{proof}
\vspace{10pt}
\begin{remark}
The condition (\ref{convexinequalitysos}) can be solved using available software packages (for example, SOSTools \cite{Prajna05}).
\end{remark}
\vspace{10pt}
Sometimes one is interested not only on stability but in some measure of performance of a stabilizing controller.
To measure performance one can consider the quadratic cost
\begin{equation}\label{cost}
J(x,u)=\sum_{k=0}^{\infty}\left(x^T(k)Qx(k)+u^T(k)Ru(k)\right)
\end{equation}
for given matrices $Q=Q^T\ge 0$ and $R=R^T> 0$.
The lower this cost is for a given controller the better is its performance.
It is possible to obtain a lower bound on the minimum value of the cost (\ref{cost}) for any asymptotically stabilizing controller by solving a convex optimization problem as stated in the next Theorem.
\vspace{10pt}
\begin{theorem}
Assume that a system model is given in the form (\ref{statespacemodelinx}) where $A(x(t))$ and $B(x(t))$ have polynomial entries.
If there exists $P=P^T$ such that
\begin{equation}\label{guaranteedcost}
W(x(t))=\left[
\begin{array}{lr}
A^T(x(t))PA(x(t))-P+Q & A^T(x(t))PB(x(t))\\
B^T(x(t)) PA(x(t)) & R+B^T(x(t))PB(x(t))
\end{array}
\right] {\rm is~SOS},~\forall~x(t)\in\RE^{n_x},
\end{equation}
then any controller that asymptotically stabilizes the system to the origin yields a cost (\ref{cost}) that is bounded below by $V(x_0)=x_0^TPx_0$ when the system trajectories start from an initial condition $x_0\in\RE^{n_x}$.
\end{theorem}
\vspace{10pt}
\begin{proof}
The condition (\ref{guaranteedcost}) implies that the matrix $W(x(t))$ is positive semidefinite for all $x(t)\in\RE^{n_x}$.
Left multiplying $W(x(t))$ by $[x^T(t)~~u^T(t)]$ and right multiplying by $[x^T(t)~~u^T(t)]^T$ and using (\ref{statespacemodelinx}) yields
\begin{equation}\label{guaranteedinequality}
x^T(t)Qx(t)+u^T(t)Ru(t)+V(x(t+1))-V(x(t))\ge 0,
\end{equation}
where $V(x(t))=x^T(t)Px(t)$.
Performing the sum of the terms on the left hand side of inequality (\ref{guaranteedinequality}) from zero to infinity yields
\begin{equation}\label{guaranteedfinalcost}
J(x,u)\ge V(x(0))-V\left(\lim_{t\to\infty}x(t)\right)=V(x_0),
\end{equation}
provided $x(t)\to 0$ as $t\to\infty$, which is guaranteed when the closed-loop system is asymptotically stable to the origin.
\end{proof}
\vspace{10pt}
\begin{remark}
For some systems in the form (\ref{statespacemodelinx}) the optimal cost to go to the origin from a given state $x$ is $V(x)=x^TPx$, where $P$ is the matrix of maximum trace that satisfies condition (\ref{guaranteedcost}).
That is the case for example when the system is linear.
When this happens, the inequality (\ref{guaranteedinequality}) becomes an equality that is called the Bellman equation \cite{Lewisbook}.
The optimal controller is then asymptotically stabilizing and is given by \cite{Lewisbook}
\begin{equation}\label{optimalcontroller}
u(t)=-\left(R+B^T(x(t))PB(x(t))\right)^{-1}B^T(x(t)) PA(x(t))x(t).
\end{equation}
A common control design heuristic is thus to find the matrix $P$ with maximum trace that satisfies condition (\ref{guaranteedcost}) and to compute the control input (\ref{optimalcontroller}), even when the value function may not be $V(x)=x^TPx$.
This decouples the design of the controller from the determination of $P$ and makes the controller synthesis problem convex at the cost of no guarantees that the controller is stabilizing.
\end{remark}
\vspace{10pt}
\begin{remark}
Note from (\ref{optimalcontroller}) that when the matrix $B^T(x(t))PB(x(t))$ is invertible for all $x(t)$ the matrix $R$ can be set to zero, leading to what is called the cheap control solution.
\end{remark}
\vspace{10pt}
\begin{remark}
The convex conditions proposed in \cite{Amatoetal2013} are sufficient conditions for stability of closed-loop nonlinear quadratic systems  of the form (\ref{statespacemodel2}) with $E(u(t))=0$ and no constant offset terms.
The conditions yield a guaranteed upper bound on the cost (\ref{cost}) when the state is constrained to a polytopic subset $\mathcal X$ of the state space.
\end{remark}

\section{Examples}\label{examples}
The following examples show the application of the theoretical results developed in the paper.
All examples were solved in cvx \cite{cvxmanual} interfaced with Matlab R2017a\footnote{Matlab is a trademark of The Mathworks Inc.} and using the convex optimization solver MOSEK \cite{mosekmanual}.
\vspace{10pt}
\begin{figure}[t] 
\centerline{ \resizebox{120mm}{!}{\includegraphics{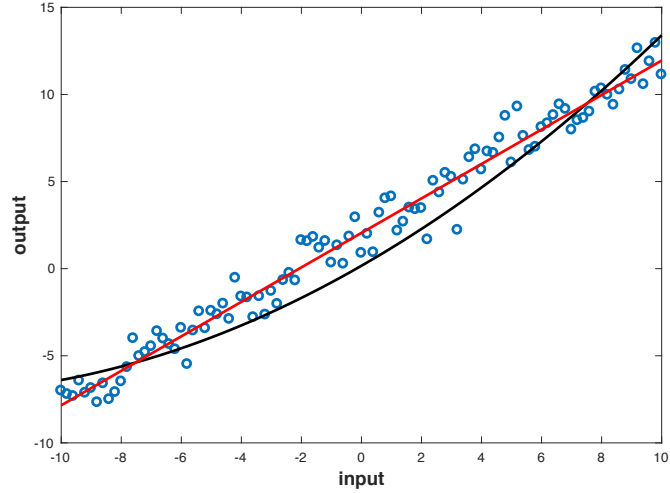}}}
\caption{Neural network approximation of $f(x)=x+2+\nu$, where $\nu\sim N(0,1)$ (black curve) and least squares approximation (red line).}
\label{regression}
\end{figure}
\begin{example}
To validate the result of Corollary \ref{regularizationresult}, we assume a single input and single output network. Since in this case the matrix $\bar Z$ in equation (\ref{networkexpression}) has three parameters to be estimated, we collected $N=101$ data points, which is much greater than the number of parameters.
The neural network training was performed for a quadratic activation function with parameters $a=0.0937, b=0.5, c=0.4688$, a regularization coefficient of $\beta=0.001$, and a quadratic norm loss function $l(\cdot)$ to approximate the function $f(x)=x+2+\nu$, where $\nu\sim N(0,1)$.
The symmetric matrix that describes the neural network is
\begin{eqnarray*}
\bar Z=\left[
\begin{array}{cc}
0.0324 & 0.5241\\
0.5241 & 0.1619
\end{array}
\right].
\end{eqnarray*}
The neural decomposition algorithm \ref{neuraldecomposition} with a tolerance of {\bf tol}$=10^{-5}$ leads to a neural network with two neurons and the following weights for the first layer
\begin{equation*}
w_{+}=\frac{1.1050}{|1.1050|}=1,\quad w_{-}=\frac{-0.9357}{|-0.9357|}=-1.
\end{equation*}
The weights for the second layer are
\begin{equation*}
\alpha_{+}=\left(1.1050\right)^2=1.221,\quad \alpha_{-}=-\left(0.9357\right)^2=-0.8755.
\end{equation*}
The neural network output is thus given by
\begin{equation*}
\hat y(x) = 1.221\sigma\left(x\right)-0.8755\sigma\left(-x\right)=.0324x^2+1.0482x+.1619.
\end{equation*}
The best-fitting line in the least squares sense is given by the equation
\begin{equation*}
y_{ls}=0.9677x+2.1302.
\end{equation*}
Figure \ref{regression} shows the neural network regression and its comparison to a best-fitting line in the least squares sense.
It is clear from the figure that the neural network regression is not an affine line as it would be the case if one were to use standard least squares line-fitting, as seen by the red line in figure \ref{regression}.
\end{example}
\vspace{10pt}
\begin{figure}[t] 
\centerline{ \resizebox{150mm}{!}{\includegraphics{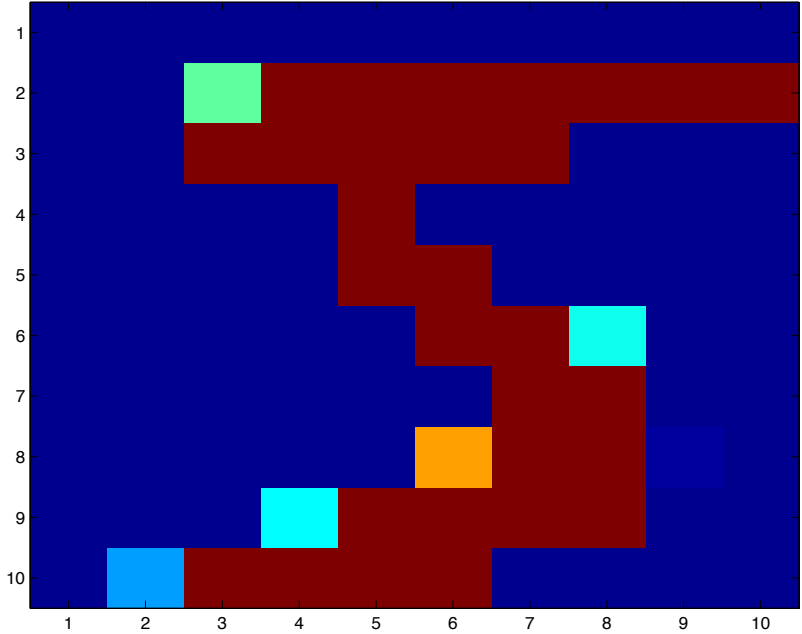}}}
\caption{Number $5$ after being downsampled.}
\label{classification}
\end{figure}

\begin{example}
In this example we will consider the classification of handwritten digits from the MNIST database \cite{MNIST}.
This database has $60,000$ digit samples for training and $10,000$ digit samples for validation.
Each digit image has $28\times28=784$ pixels.
The first step before the classification of the digits is to downsample the images.
We have chosen to only use the rows and columns $5$ to $24$ of each image and then downsample this smaller image so that it would only have $10\times10=100$ pixels.
This downsampled image has therefore only $12.76\%$ of the pixels of the original image.
Figure \ref{classification} shows one of the downsampled images for the number $5$.
The reason to perform the downsampling operation is because the neural network parameters will be the entries of the $(n+1)\times(n+1)$ symmetric matrix in expression (\ref{networkexpression}), where $n$ is the number of pixels.
The larger the number of pixels the larger this matrix will be and therefore the slower will be the training of the neural network.
The training was done for a quadratic activation function with parameters $a=0.0937, b=0.5, c=0.4688$ using a regularization coefficient of $\beta=590$ and an infinity norm loss function $l(\cdot)$.
The results of the neural network training and digit classification (testing) for different batches of $10\times10$ images are presented in table \ref{digitrecognitionresults}.
The percentages of testing are calculated based on the number of correctly classified digits out of the $10,000$ validation set that was not used to train the neural network.
The training time was measured on a MacBookAir $1.6$GHz with $16$GB memory.
Note that although a success rate of $93.31\%$ is smaller than the typical $96-98\%$ of a feedforward (non-CNN) neural network using deep learning \cite{Allgoweretall2022}, only $5.37\%$ of the total number of training images were used to achieve a $93.31\%$ success rate.
This shows that for this particular case a quadratic feedforward neural network is promising in terms of obtaining very good results with only a fraction of the total data available for training.
Additionally, the worst case Lipschitz constant for all outputs $L_n^{max}$ with $n=10^2$ was computed for the neural network with highest success rate using the result of Theorem \ref{Lipschitz}.
The Lipschitz constant is $L_{100}^{max}=3.7\times 10^{-5}$.
This small value means that the quadratic neural network has small sensitivity to changes in the input and is therefore robust to noisy inputs.
\begin{table}[h]
	\begin{center}
	\caption{Digit classification results}
	\label{digitrecognitionresults}
	\begin{tabular}{|c|c|c|}
\hline
Training Batch Size & Training Time (hours) & Success Rate ($\%$)
\\ \hline
		\hline
		100	&	0.21		&	70.89		
		\\ \hline
		250	&	0.32	&	82.22
		\\ \hline
		500	&	0.44	&	86.54
		\\ \hline
		1000		&	1.50		&	89.75
		\\ \hline
		1400		&	2.50		&	90.56
		\\ \hline
		1600		&	3.40		&	91.00
		\\ \hline
		1920		&	4.30		&	91.53
		\\ \hline
		3000		&	8.50		&	92.80
		\\ \hline
		3225		&	19.3	&	93.31
		\\ \hline
	\end{tabular}
	\end{center}
\end{table}
\end{example}
\vspace{10pt}

\begin{example}
This example considers the problem of system identification of a flexible robot arm whose input-output data with reference [$96-009$] is available at the website of the Database for the Identification of Systems (DaISy) at the university of KU Leuven \cite{Daisy}.
The input is the measured reaction torque of the arm structure on the ground and the output is the acceleration of the flexible arm.
The applied input was a periodic sine sweep for which an output consisting of $N=1024$ data points was collected.
The input and output matrices for the neural network training are the ones from (\ref{systemidmatrices}) with $n=1$.
The training was done for a quadratic activation function with parameters $a=0.0937, b=0.5, c=0.4688,$ using a regularization coefficient of  $\beta=0.01$ and an infinity norm loss function $l(\cdot)$. 
Only the first $122$ data points were used to train the neural network, which represents approximately $12\%$ of all data points.
The resulting (one state $x(t)=y(t)$) model is described by
\begin{equation*}
y(t+1)=\left[
\begin{array}{c}
u(t)\\
y(t)\\
1
\end{array}
\right]^T
\bar Z\left[
\begin{array}{c}
u(t)\\
y(t)\\
1
\end{array}
\right],
\end{equation*}
where
\begin{eqnarray*}
\bar Z=\left[
\begin{array}{rrr}
-0.0989 & 0.5030 &-0.1682\\
0.5030 & 0.1599 & 1.8265\\
-0.1682 & 1.8265 & 0.0610
\end{array}
\right].
\end{eqnarray*}

The model prediction is compared with the data in figure \ref{flexiblearm}.
\begin{figure}[t] 
\centerline{ \resizebox{120mm}{!}{\includegraphics{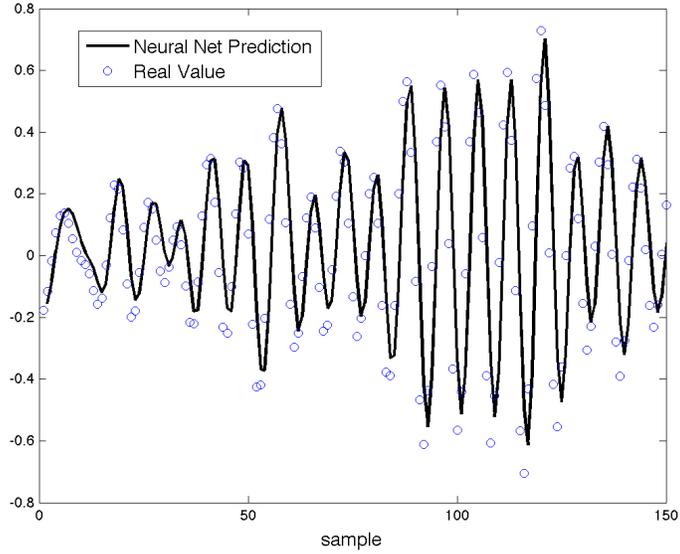}}}
\caption{System identification (black curve) and data points (blue circles).}
\label{flexiblearm}
\end{figure}
\end{example}
\vspace{10pt}

\begin{example}\label{firstordersystem}
Consider the quadratic single state control system
\begin{equation}\label{simplemodel}
x(t+1) = x^2(t) +x(t)u(t) + u^2(t) - 1,
\end{equation}
which is in the form (\ref{statespacemodel2}) with
\begin{eqnarray*}
\bar A(x(t))=\left[
\begin{array}{lr}
x(t) & -1\\
0 & 1
\end{array}
\right],\quad
\bar B(x(t))=\left[
\begin{array}{c}
x(t)\\
0
\end{array}
\right],\quad
E(u(t))=\left[
\begin{array}{c}
u(t)\\
0
\end{array}
\right].
\end{eqnarray*}
In steady state one can write $x(t+1)=x(t)=x_*$. For a desired steady state of $x_*=0$, equation (\ref{simplemodel}) becomes
\begin{equation*}
u_*^2 = 1,
\end{equation*}
which has the solutions $u_*=1$ and $u_*=-1$.
If the constant offset term $-1$ was not present in the model (\ref{simplemodel}) then $u_*=0$ would be the single steady state solution for the control input.
Notice that the system (\ref{simplemodel}) can also be written in the form (\ref{statespacemodel}) as
\begin{equation}
x(t+1)=\left[
\begin{array}{c}
u(t)\\
x(t)\\
1
\end{array}
\right]^T\left[
\begin{array}{ccr}
1 & 0.5 & 0\\
0.5 & 1 & 0\\
0 & 0 & -1
\end{array}
\right]\left[
\begin{array}{c}
u(t)\\
x(t)\\
1
\end{array}
\right],
\end{equation}
which could represent the output of a quadratic neural network if, for example, the parameters of the network activation function were $a=1, b=0, c=-0.5$, with network inputs $u(t)$ and $x(t)$.\done
\end{example}
\vspace{10pt}

\begin{example}\label{simplerexample}
Consider now the single state system obtained from (\ref{firstordersystem}) by removing the offset term and the 	quadratic term in $u$ written as
\begin{equation}\label{simplermodel}
x(t+1) = x^2(t) +x(t)u(t).
\end{equation}
Assuming that the desired steady state setpoint is $x_*=0$ and solving (\ref{simplermodel}) for the steady state control input one finds that any finite value of $u_*$ leads to a steady state value of $x_*=0$.
Therefore, one can choose $u_*=0$.
Assuming a constant state feedback gain then the control input (\ref{control}) becomes $u(t)=Kx(t)$.
Replacing this input into (\ref{simplermodel}) yields the closed-loop dynamics\begin{equation}\label{simplerclosedloop}
\bar x(t+1) = [(1+K)x^2(t)~~1]^T=A_{cl}(x(t),K)\bar x(t),
\end{equation}
where
\begin{eqnarray*}
A_{cl}(x(t),K)=\left[
\begin{array}{cc}
(1+K)x(t) & 0\\
0 & 1
\end{array}
\right].
\end{eqnarray*}
The candidate Lyapunov function (\ref{Lyapunovfunction}) becomes $V(x)=px^2$ with $p>0$.
The condition (\ref{newdecreasingcondition}) is then written as 
\begin{eqnarray*}
\left[
\begin{array}{cc}
p(1+K)^2x^2(t)-p & 0\\
0 & 0
\end{array}
\right]\le 0,~\forall~x(t),
\end{eqnarray*}
which is equivalent to $p(1+K)^2x^2(t)-p\le0,~\forall x(t)$.
This constraint is not linear in the unknowns $p$ and $K$.
It is however easy to see that the choice $K=-1$ leads to a global solution independently of $x(t)$ for any $p>0$.
The control input $u(t)=-x(t)$  takes the trajectories to zero in one time step and is called a dead beat controller. The asymptotic stability of the closed-loop system is global because $V(x)$ is radially unbounded for any $p>0$.
\done
\end{example}
\vspace{10pt}
\begin{example}
Consider a model of a quadrotor flying at a constant altitude in the positive $X$ direction (see Fig.~\ref{fig:quad_schematic})
\begin{equation}
\label{quadrotor_nonlinear}
	\begin{bmatrix}
		\dot X \\ \dot{V_X}
	\end{bmatrix}
	=
	\begin{bmatrix}
		0 & 1 \\ 0 & 0
	\end{bmatrix}
	\begin{bmatrix}
		X \\ V_X
	\end{bmatrix}
	+
	\begin{bmatrix}
		0 \\ g
	\end{bmatrix}
	u
	-
	\begin{bmatrix}
		0 \\ d
	\end{bmatrix}
	V_X^2,
\end{equation}
where $X$ and $V_X$ represent the position and the velocity of the quadrotor, respectively, $u=\tan(\theta)$ is the control input where $\theta$ is the pitch angle of the quadrotor (positive counter clockwise), $g$ is the acceleration of gravity, and $d=\frac{\rho SC_D}{2m}$ is the term related to the drag force where $\rho$ is the air density, $S$ is a characteristic surface, $m$ is the mass of the quadrotor, and $C_D$ is the coefficient of drag.
\begin{figure}[t]
\centering \includegraphics[width=9cm]{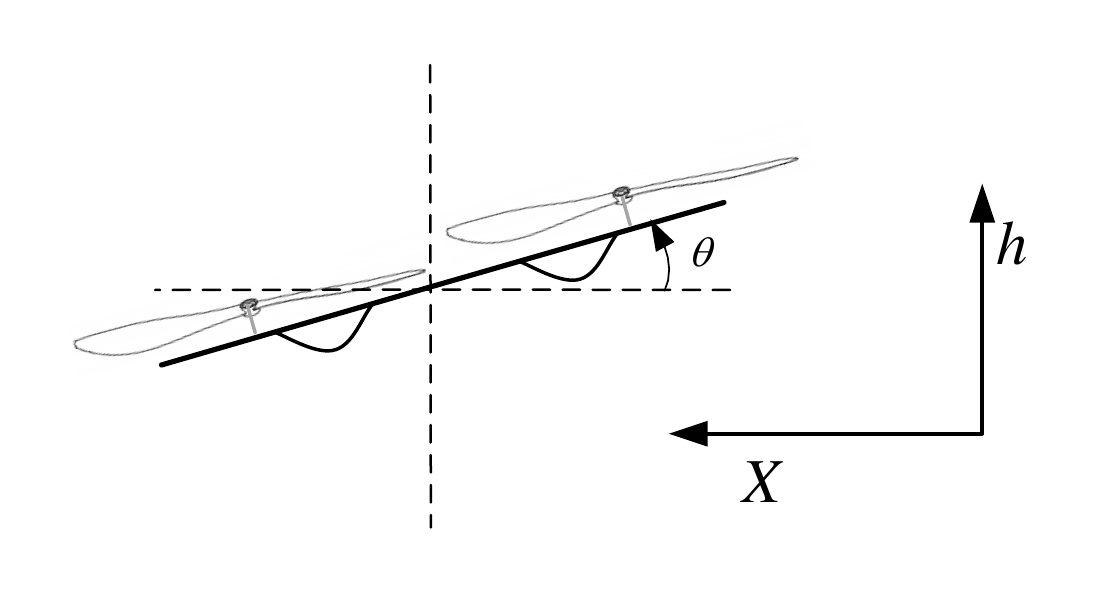}
	\caption{Quadrotor model in forward flight at constant altitude.}
	\label{fig:quad_schematic}
\end{figure}
The control objective is to take the system to the origin from any initial position with a negative $X$ coordinate.
The model (\ref{quadrotor_nonlinear}) is in continuous-time and will be discretized by approximating the derivative as
\begin{equation}\label{derivativeapproximation}
\dot X \approx \frac{X(t+T)-X(t)}{T}
\end{equation}
where $T$ is the sampling time.
The same approximation will be used for $\dot V_X$.
The approximate discrete-time model model can be written in the form (\ref{statespacemodelinx}) where $x(t)=[X(t)~~V_X(t)]^T$ and
\begin{eqnarray*}
A(x(t))=\left[
\begin{array}{lc}
1\quad & T\\
0\quad & 1-TdV_X(t)
\end{array}
\right],\quad
B(x(t))=\left[
\begin{array}{c}
0\\
Tg
\end{array}
\right].
\end{eqnarray*}
\end{example}
The parameters of the system are $T=0.102s, g=9.8ms^{-2}, \rho=1.097kgm^{-3}, S=1m^2,C_D=0.05,m=1.2kg$, which yield $Td=0.0023sm^{-1}$ and $Tg=1ms^{-1}$.
Finding the matrix $P$ with minimum condition number that satisfies condition (\ref{convexinequality}) with $\epsilon=0.1$ and $L(x)$ a first order polynomial yields
\begin{eqnarray*}
P=\begin{bmatrix}
1.4589 & -1.6008\\-1.6008 & 2.6636
\end{bmatrix},
\end{eqnarray*}
which represents a quadratic neural network with $ca^{-1}=(2.6636)(1.4589 )^{-1}=1.8258$.
The controller is given by
\begin{eqnarray*}
u(t) = -1.1556X(t)-1.1771V_X(t)+0.0023V_X^2(t).
\end{eqnarray*}
Note that the controller cancels the term in $V_X^2(t)$ from the open-loop system and yields a linear closed-loop system with stable eigenvalues $\lambda_1=0.8895, \lambda_2=-0.0666$.

Finding the matrix $P$ with maximum trace that satisfies condition (\ref{guaranteedcost}) for $Q=I$ and $R=2.2\times10^{-16}$ using SoSTools~\cite{Prajna05} yields
\begin{eqnarray*}
P=\begin{bmatrix}
11.3167 & 1.0523\\1.0523 & 1.1073
\end{bmatrix},
\end{eqnarray*}
which represents a quadratic neural network with $ca^{-1}=0.0979$.
Using equation (\ref{optimalcontroller}) leads to the controller
\begin{eqnarray*}
u(t) = -0.9503X(t)-1.097V_X(t)+0.0023V_X^2(t).
\end{eqnarray*}
Note that this controller also cancels the term in $V_X^2(t)$ from the open-loop system and yields a linear closed-loop system with stable eigenvalues $\lambda_1=0.9031, \lambda_2=-0.0001$.
The control input can be rewritten as the quadratic form
\begin{equation*}
u(t)=
\begin{bmatrix}
X(t)\\
V_X(t)\\
1
\end{bmatrix}^T
\begin{bmatrix}
0 & 0 & -0.47515\\
0 & 0.0023 & -0.5485\\
-0.47515 & -0.5485 & 0
\end{bmatrix}
\begin{bmatrix}
X(t)\\
V_X(t)\\
1
\end{bmatrix},
\end{equation*}
and therefore it can be implemented by a quadratic neural network with $c=0$.

\section{Conclusions}
This paper addressed the analysis and design of quadratic neural networks for regression, classification, system identification, and control of dynamical systems.
All problems were formulated as a convex optimization, which can be efficiently solved with polynomial-time algorithms.
The following conclusions can be taken from this work:
\vspace{10pt}
\begin{itemize}
\item Quadratic neural networks have the advantage that the training of the weights leads to the global optimum value, the number of neurons and network architecture is a by-product of the training, and the input-output are related by a quadratic form leading to an analytical bound to the sensitivity of the output to small changes in the input.
\item Since the length of the quadratic parametric description of the network is related to the square of the length of the input $x$, the neural network training time will rapidly grow with the input length.
It would be important in future research to develop dedicated convex optimization routines that could exploit the sparsity in the data and possible symmetries in the architecture of quadratic neural networks.
\item The way in which the training is coded is an important factor to make the training faster. It is advantageous to use the Kronecker product to vectorize the quadratic form taking into account all monomial terms in $xx^T$. Avoiding repetition of monomials will imply less computational time during the training. For example, if $x=[x_1~x_2]^T$ then the monomials of $xx^T$ are $(x_1^2,x_1x_2,x_2x_1,x_2^2)$ and this set should be replaced by the smaller set $(x_1^2,x_1x_2,x_2^2)$ to avoid repetitions.
\item From the examples in this paper it appears that quadratic neural networks work extremely well with only a small fraction (5\%-20\%) of the available training data. This empirical evidence is very promising but must be confirmed theoretically in future studies. 
\item Neural networks have not yet found a vast application in safety-critical control systems such as autonomous driving due to the absence of stability certificates for the closed-loop system. By casting the controller design as a convex program yielding a Lyapunov function as a stability proof guarantee, this paper has opened the door for more applications of neural networks in safety critical applications. To the best of our knowledge it is the first time that controller synthesis for neural networks with a stability certificate as a deliverable has been formulated and solved as a convex optimization problem.
\end{itemize}

\section{Acknowledgements}
The authors would like to thank Burak Bartan and Mert Pilanci from Stanford University for the answers to all our questions about their work on quadratic neural networks presented in reference \cite{BartanPilanci2021}. In particular, we would like to thank Burak Bartan for making his code available in a Python notebook.
The authors would also like to thank the Natural Sciences and Engineering Research Council (NSERC) for funding this research.

\bibliographystyle{IEEEtran}
\bibliography{nnbibliography}

\begin{thebibliography}{10}
\providecommand{\url}[1]{#1}
\csname url@samestyle\endcsname
\providecommand{\newblock}{\relax}
\providecommand{\bibinfo}[2]{#2}
\providecommand{\BIBentrySTDinterwordspacing}{\spaceskip=0pt\relax}
\providecommand{\BIBentryALTinterwordstretchfactor}{4}
\providecommand{\BIBentryALTinterwordspacing}{\spaceskip=\fontdimen2\font plus
\BIBentryALTinterwordstretchfactor\fontdimen3\font minus
  \fontdimen4\font\relax}
\providecommand{\BIBforeignlanguage}[2]{{%
\expandafter\ifx\csname l@#1\endcsname\relax
\typeout{** WARNING: IEEEtran.bst: No hyphenation pattern has been}%
\typeout{** loaded for the language `#1'. Using the pattern for}%
\typeout{** the default language instead.}%
\else
\language=\csname l@#1\endcsname
\fi
#2}}
\providecommand{\BIBdecl}{\relax}
\BIBdecl

\bibitem{McCullochPitts1943}
W.~S. McCulloch and W.~Pitts, ``A logical calculus of the ideas immanent in
  nervous activity,'' \emph{Bulletin of Mathematical Biophysics}, vol.~5, pp.
  115--133, 1943.

\bibitem{Pavonetal2021}
H.~Dai, B.~Landry, L.~Yand, M.~Pavone, and R.~Tedrake, ``Lyapunov-stable
  neural-network control,'' in \emph{Proceedings of Robotics: Science and
  Systems 2021}, Held Virtually, July 12-16 2021.

\bibitem{Huang1998}
G.-B. Huang and H.~Babri, ``Upper bounds on the number of hidden neurons in
  feedforward networks with arbitrary bounded nonlinear activation functions,''
  \emph{IEEE Transactions on Neural Networks}, vol.~9, no.~1, pp. 224--229,
  1998.

\bibitem{Cheridito2021}
P.~Cheridito, A.~Jentzen, and F.~Rossmannek, ``Efficient approximation of
  high-dimensional functions with neural networks,'' \emph{IEEE Transactions on
  Neural Networks and Learning Systems}, pp. 1--15, 2021.

\bibitem{Luo2021}
Y.~Luo, J.~Lü, X.~Jiang, and B.~Zhang, ``Learning from architectural
  redundancy: Enhanced deep supervision in deep multipath encoder-decoder
  networks,'' \emph{IEEE Transactions on Neural Networks and Learning Systems},
  pp. 1--14, 2021.

\bibitem{Fazlyab2019}
M.~Fazlyab, A.~Robey, H.~Hassani, M.~Morari, and G.~J. Pappas, ``Efficient and
  accurate estimation of {L}ipschitz constants for deep neural networks,'' in
  \emph{Proceedings of the 33rd International Conference on Neural Information
  Processing Systems (NIPS)}, December 2019, pp. 11\,427--11\,438.

\bibitem{Allgoweretall2022}
P.~Pauli, A.~Koch, J.~Berberich, P.~Kohler, and F.~Allg\"ower, ``Training
  robust neural networks using {L}ipschitz bounds,'' \emph{IEEE Control Systems
  Letters}, vol.~6, pp. 121--126, 2022.

\bibitem{Bengioetal2016}
I.~Goodfellow, Y.~Bengio, and A.~Courville, \emph{Deep Learning}.\hskip 1em
  plus 0.5em minus 0.4em\relax MIT Press, 2016,
  \url{http://www.deeplearningbook.org}.

\bibitem{BartanPilanci2021}
B.~Bartan and M.~Pilanci, ``Neural spectrahedra and semidefinite lifts: Global
  convex optimization of polynomial activation neural networks in fully
  polynomial-time,'' \emph{arxiv.org}, 2021.

\bibitem{Algebrabook}
C.~Audet, P.~Hansen, and G.~S. Editors, \emph{Essays and Surveys in Global
  Optimization}.\hskip 1em plus 0.5em minus 0.4em\relax Springer, 2005.

\bibitem{Wolkowicz-Styan1980}
H.~Wolkowicz and G.~P.~H. Styan, ``Bounds for eigenvalues using traces,''
  \emph{Linear Algebra and its Applications}, vol.~29, pp. 471--506, 1980.

\bibitem{StochasticBook}
D.~A. Kendrick, \emph{Stochastic Control for Economic Models}, 2nd~ed.\hskip
  1em plus 0.5em minus 0.4em\relax McGraw-Hill Inc., 2002.

\bibitem{Ljungbook}
L.~Ljung, \emph{System Identification: Theory for the User}.\hskip 1em plus
  0.5em minus 0.4em\relax Prentice Hall, 1999.

\bibitem{Prajna05}
S.~Prajna, A.~Papachristodoulou, P.~Seiler, and P.~A. Parrilo, \emph{{SOSTOOLS
  and Its Control Applications}}.\hskip 1em plus 0.5em minus 0.4em\relax
  Springer-Verlag, 2005, pp. 273--292.

\bibitem{Lewisbook}
F.~L. Lewis, D.~L. Vrabie, and V.~L. Syrmos, \emph{Optimal Control},
  3rd~ed.\hskip 1em plus 0.5em minus 0.4em\relax John Wiley {\&} Sons, Inc.,
  2012.

\bibitem{Amatoetal2013}
F.~Amato, R.~Ambrosino, M.~Ariola, and A.~Merola, ``Domain of attraction and
  guaranteed cost control for non-linear quadratic systems. part 2: controller
  design,'' \emph{IET Control Theory Applications}, vol.~7, no.~4, pp.
  565--572, 2013.

\bibitem{cvxmanual}
M.~C. Grant and S.~Boyd, \emph{The CVX User's Guide Release 2.1}, CVX Research,
  Inc., December 2017.

\bibitem{mosekmanual}
\emph{MOSEK Modeling Cookbook Release 3.2.3}, November 2021.

\bibitem{MNIST}
L.~Deng, ``The {MNIST} database of handwrittten digit images for machine
  learning research,'' \emph{IEEE Signal Processing Magazine}, vol.~29, no.~6,
  pp. 141--142, 2012.

\bibitem{Daisy}
B.~D. Moor, P.~D. Gersem, B.~D. Schutter, and W.~Favoreel, ``Daisy: A database
  for identification of systems,'' \emph{Journal A}, vol.~38, no.~3, pp. 4--5,
  1997.

\end{thebibliography}
\end{document}